\documentclass[conference]{IEEEtran}
\IEEEoverridecommandlockouts
\usepackage{cite}
\usepackage{amsmath,amssymb,amsfonts}
\usepackage{algorithmic}
\usepackage{graphicx}
\usepackage{textcomp}
\usepackage{xcolor}

\def\BibTeX{{\rm B\kern-.05em{\sc i\kern-.025em b}\kern-.08em
    T\kern-.1667em\lower.7ex\hbox{E}\kern-.125emX}}
\begin{document}

\title{Banking on Feedback: Text Analysis of Mobile Banking iOS and Google App Reviews}

\author{\IEEEauthorblockN{1\textsuperscript{st} Yekta Amirkhalili}
\IEEEauthorblockA{\textit{Department of Management Engineering} \\
\textit{University of Waterloo}\\
ON, Canada \\
yamirkha@uwaterloo.ca}
\and
\IEEEauthorblockN{2\textsuperscript{nd} Ho Yi Wong}
\IEEEauthorblockA{\textit{Department of Management Engineering} \\
\textit{University of Waterloo}\\
ON, Canada \\
hy3wong@uwaterloo.ca}
}

\maketitle

\begin{abstract}
The rapid growth of mobile banking (m-banking), especially after the COVID-19 pandemic, has reshaped the financial sector.  
This study analyzes consumer reviews of m-banking apps from five major Canadian banks, collected from Google Play and iOS App stores.  
Sentiment analysis and topic modeling classify reviews as positive, neutral, or negative, highlighting user preferences and areas for improvement.  
Data pre-processing was performed with NLTK, a Python language processing tool, and topic modeling used Latent Dirichlet Allocation (LDA).  
Sentiment analysis compared methods, with Long Short-Term Memory (LSTM) achieving 82\% accuracy for iOS reviews and Multinomial Naive Bayes 77\% for Google Play.  
Positive reviews praised usability, reliability, and features, while negative reviews identified login issues, glitches, and dissatisfaction with updates.  
This is the first study to analyze both iOS and Google Play m-banking app reviews, offering insights into app strengths and weaknesses.  
Findings underscore the importance of user-friendly designs, stable updates, and better customer service. Advanced text analytics provide actionable recommendations for improving user satisfaction and experience.

\end{abstract}


\section{Introduction}
\label{introduction}
The COVID-19 pandemic has significantly impacted the banking industry, accelerating the adoption of digital banking \cite{b1}.
Mobile banking (m-banking), which allows banking via mobile devices, is now the leading banking method in Canada as of $2024$ \cite{b4}. 
Although m-banking apps are not the sole factor in choosing a bank, poor app performance can lead to dissatisfaction and customer attrition \cite{b5}. 
This study analyzes consumer feedback to identify improvement areas and effective strategies for m-banking apps, aiming to enhance user experiences.
App reviews were chosen as the data source after a Scopus search\footnote{The following search was used: TITLE-ABS-KEY(("mobile banking" OR "mbanking" OR "m-banking") AND ("text analytics" OR "text analysis" OR "consumer review" OR "customer review" OR "review analysis")).}, which found only $13$ relevant studies, all published post-2021, with most from $2023–2024$.
This indicates m-banking review analysis is an emerging research area.
Reviews for five major Canadian banking apps (CIBC, TD, RBC, Scotiabank, and BMO) were collected from the Google Play Store and iOS App Store\footnote{We consider only the main (banking) app for each.}. 
The selection of these specific banks was straightforward -- they were the first five results when searching for "bank" in the search bars of both app stores.

To the best of our knowledge, no prior research has analyzed app reviews from both platforms in the context of m-banking, particularly in Canada.
This makes our study unique in its scope and contribution. 
By analyzing user reviews from Google Play and the iOS App Store, we aim to uncover the characteristics of m-banking apps that resonate most with users and identify the areas that need improvement. 
For brevity, we will use m-banking apps to refer to the specific 5 Canadian m-banking apps chosen onward.
Our paper answers three research questions: 
\begin{enumerate}
    \item \textit{RQ1}: What are m-banking users discussing and focusing on in general? 
    \item \textit{RQ2}: What are the main strengths of m-banking apps that users frequently praise? 
    \item \textit{RQ3}: What are the main problems of m-banking apps that users frequently complain about?
\end{enumerate}
Understanding these insights will help reveal what banks are doing right and where they are falling short, offering valuable guidance for improving m-banking services in Canada.
To address these questions, we use web scrapers to gather reviews from the app stores.
We analyze these reviews to identify the main topics of discussion, as well as the most frequent words and phrases (\textit{RQ1}).
Next, we categorize the reviews into three sentiment types: positive, neutral, and negative.
For each sentiment category, we identify the most frequent words and key topics.
This allows us to understand the general discussions about the apps, and to specifically examine what satisfied users are talking about (\textit{RQ2)}, and what the dissatisfied users are concerned with (\textit{RQ3}).
For the remainder of the paper, we analyze the iOS App and Google Play Store datasets separately.

\section{Data and Methods}
\label{Data ana Methods}
In this section, we will present the data and methodology for the paper. 
The information provided on the data is true as of the collection date of October $29$, $2024$. 
The data analysis is conducted in both Python (version $3.9$) and R. 

\subsection{Data}
\label{sub-data}
The first dataset includes app features for the five banks, providing an overview of the m-banking apps studied. 
In the \textbf{iOS App Store}, the median category ranking (finance) is $5$, with an average rating of $4.56 \pm 0.23$ (highest: $4.8$). 
The average number of user ratings is $236,380$, with a large standard deviation of $301,349$, showing significant popularity differences.
Number of ratings range from $21,000$ to $761,500$. 
Although exploring these is interesting, it is beyond our study's scope; we do not wish to compare banks or these apps. 
For reviews, $11,980$ were collected from the iOS App Store, excluding $823$ non-English entries\footnote{Web scraping tools limited iOS reviews to $4,000$ per app, unlike Google Play, which allowed larger datasets.}.
The average rating for users who left reviews is $2.55$, notably lower than the overall app rating of $4.56$. 
This discrepancy aligns with research suggesting extreme experiences drive reviews \cite{b6}. 
That is, consumers with very bad or very good experiences are more likely to provide online reviews.

An interesting observation is that the app with the smallest number of ratings has the highest number of reviews, indicating, a larger proportion of users for this particular app felt the need to leave a review. 
Review text lengths ranged from $7$ to $2,926$ characters, averaging $200$. 
On the \textbf{Google Play Store}, the median ranking (Finance) is $4$, with the top rank at $2$.
The average app rating is $3.84 \pm 0.8$, with the highest-rated app at $4.6$. 
Apps received an average of $91,340$ ratings, with a wide range from $43,200$ to $180,000$ (SD = $56,441$).
From $107,026$ collected reviews, $17,683$ non-English entries were removed, leaving $89,343$ for analysis.
Written reviews had an average rating of $3.16$, slightly lower than the overall $3.84$. Review lengths averaged $114$ characters, totaling $2,167$ characters.

The dataset is cleaned by converting text to lowercase, removing stop words\footnote{The stop word list from the NLTK package is extended with variations of "bank," "app," and bank names.}, eliminating punctuation, and applying lemmatization/stemming.
Reviews are tokenized into unigrams, bigrams, and trigrams (single words, word pairs, and three-word groups).
To analyze the data, we identify the most frequent words using two methods: 1) simple word counting, based on the number of occurrences across the corpus, and 2) Gensim (version $4.1.2$) for Term Frequency-Inverse Document Frequency (TF-IDF)\footnote{https://radimrehurek.com/gensim/models/tfidfmodel.html} to measure word importance, based on the following formulation:
\begin{equation}
    \label{tfidf}
    \begin{split}
        & TF(t, d) = \frac{\text{count of $t$ in $d$}}{\text{number of words in $d$}}\\
        & df(t) = \text{number of documents containing term t}\\
        & IDF(t) = log(\frac{N}{df(t)})\\
        & \text{TF-IDF}(t, d) = TF(t, d) \times IDF(t) \\
    \end{split}
\end{equation}
Where $t$ is the term or token, $d$ is the document, $TF$ calculates the frequency of a single term across a document, $df(t)$ is the document frequency for a term $t$, and $IDF$ calculates how frequent a word is across the entire corpus. 

For topic modeling, we use Latent Dirichlet Allocation (LDA) model. 
The idea of LDA is that documents are random shuffles of topics which are unobserved (latent), but can be described by a distribution over words \cite{b7}. 
To perform topic modeling using LDA, the algorithm needs to know the number of topics beforehand. 
To find the optimal number of topic across the different corpora (based on different n-grams), we try $11$ different models from $|T| = 5$ to $|T| = 15$ where $|T|$ is the number of topics. 
The choice of these values ($5$ to $15$) is based on the literature. 
After careful review of the literature, we identify ten of the most frequently cited topics in consumer reviews, which are: 

(1) \textbf{Aspects (Features)}, reviews that mention specific product attributes like price \cite{titov_modeling_2008, he_voice_2020},
(2) \textbf{Quality (Service/Product)} \cite{park_structural_2018},
(3) \textbf{Sentiment (Satisfaction/Dissatisfaction)}, reviews that discuss happiness, frustration, trust, and other positive or negative emotional responses \cite{yang_sdtm_2020},
(4) \textbf{Functionality (Technical Aspects/Performance)} \cite{ChannelSignalPaulKirwin2016},
(5) \textbf{Aesthetics (Design)} \cite{joireman_editorial_2016, ChannelSignalPaulKirwin2016},
(6) \textbf{Problems (Issues/Bugs)} \cite{ahmad_analyzing_2017, igarashi_interpretable_2020},
(7) \textbf{Usefulness} \cite{joireman_editorial_2016, ChannelSignalPaulKirwin2016}, 
(8) \textbf{Ease of Use (Usability/User Experience)} \cite{ChannelSignalPaulKirwin2016},
(9) \textbf{Content} \cite{wang_semantic_2022, PowerReviews22, ignitingBusiness21}, and (10) \textbf{Customer Service Experience} \cite{Peekage24, ignitingBusiness21, park_structural_2018}

However, we observe that some of the topics addressed are subtopics of others.
For example, \textit{Design/Aesthetics} are one \textit{Aspect} or \textit{Feature} of a product. 
Additionally, there are various topics that are domain specific \cite{tushev_domain_specific_2022}. 
Therefore, there might be additional topics not mentioned here depending on different domain. 
Consequently, $10$ topics may either be too many or too few topics to consider. 
We decide to train our topic model with $5$ to $15$ number of topics to account for the fact that we may see domain specific topics as well as the fact that we might potentially want to combine topics together.  
We run the $11$ variations of models for $3$ n--grams: uni--grams, bi--grams and tri--grams. 
In total, $33$ models are trained and the results are compared. 

We compare the models based on perplexity and coherence scores, where lower perplexity and higher coherence measures are better.
The formula for perplexity $p$ and coherence $c$ are as follows: 
\begin{equation}
    \label{pandc}
    \begin{split}
        & p = e^{- \frac{1}{N} \sum_{d = 1}^{D}\sum_{w \in d} log P(w|d)}\\
        & c(T,V) = \frac{1}{|T|} \sum_{t \in T} \frac{1}{\ \binom{|t|}{2}} \sum_{(w_i, w_j) \in t} score(w_i, w_j)\\
    \end{split}
\end{equation}
Where in the perplexity formula, $N$ is the number of words in the corpus, $D$ is the number of documents, $w$ is a single word in a document, and $P(w|d)$ is the probability that word $w$ belongs to document $d$.  
In the coherence formula, $T$ is the set of topics, $t$ is a single topic, $(w_i, w_j)$ is a word pair, and $score(w_i, w_j)$ measures the how often two words $w_i$ and $w_j$ appear together in the corpus.  

Next, we conduct sentiment analysis to categorize reviews into positive, negative and neutral. 
First, we manually label the data by comparing each review's rating with the average rating. 
If a review's rating is at least one standard deviation higher than the average, it is labeled as ``positive'', if it is at least one standard deviation lower than the average rating, it is labeled as ``negative'' and if it is within one standard deviation of the average rating it is labeled as ``neutral''. 
While this may seem arbitrary, we believe that the trends observed in the data with the average rating almost at exactly the mid-point provides support for this segmentation for iOS data and the large number of reviews balances things for the Google data.
Additionally, the authors verified the accuracy of the ratings by reading several reviews, which were chosen randomly. 

Next, we use four methods for sentiment analysis, two specific sentiment analysis Python modules and two machine learning algorithms: TextBlob, VADER (NLTK version $3.7$), multinomial Na\"ive Bayes and Long Short-Term Memory (tensorflow version $2.10.0$). 
We split the data into training (80\%) and test sets (20\%).
We train the models on the training data and compare the results with the test set (manually labeled). 
Both sentiment analysis approaches give each review a score.
In TextBlob, this score is called the polarity, which is a value from $-1$ to $1$ with values closer to $-1$ indicating a more negative and $+1$, a positive sentiment. 
Subjectivity which measures how factual vs how subjective a statement is is ignored as all reviews are based on opinions. 
To convert this score into labels, we break the interval of $[-1, +1]$ into $3$ segments and consider any score lower and equal to $-1/3$ to be a negative label, higher and equal than/to $1/3$ to be positive and neutral otherwise. 

For VADER, which is an NLTK dictionary used to do sentiment analysis, the algorithm produces a dictionary with scores for negative, positive and neutral possibilities and a compound score which is a value in $[-1, 1]$ similar to polarity from TextBlob. 
However, to decide on a label based on the results from this sentiment analysis, a threshold is considered and is compared with compound values. 
Upon review of the industry practices, it seems that a threshold of $0.05$ works best and this aligns with our experience as well (default setting according to NLTK implementation). 
Values less than $-0.05$ compound score are considered negative labeled, higher than $0.05$ as positive labeled and anything else a neutral label. 

Classification algorithms use the manual labels generated to train and learn patterns to categorize data (only based on text in the reviews and not ratings).
For neural networks, we can add a word embedding layer which converts words to embeddings such that semantically similar words have the same embeddings.
Multinomial Na\"ive Bayes is our classification algorithm with multiple classes (3), and LSTM is the recurrent neural network used for the same task (to label reviews as positive, negative or neutral). 
The approach with the highest accuracy is chosen for sentiment analysis and review labels. 

LSTM model works the best across the board for iOS data, and is chosen for labeling.
For the iOS data, the entire dataset was used in sentiment analysis and labels were generated through the following LSTM model: $5$ layers with embedding layer ($40$\% dropout rate), LSTM layer (recurrent dropout rate of $20$\%), hidden deep layer (LeakyReLU activation), and the last layer with softmax activation, Adam optimizer with Categorical CrossEntropy loss function. 
For the Google dataset, the Multinomial Naive Bayes (MNB) model delivered the best performance.
The process involved pre-processing the text data to create a word count matrix using the CountVectorizer function in Python, and training the MNB classifier on labeled review data.

\section{Results}
\label{Results}
In this section, we present the results of the study in detail for each task and discuss the implications of the results. 

\subsection{Topic Modeling}
\label{topicmodeling}
A simple counting of words across the entire corpus of iOS reviews returns the top 10 frequent words as follows: update, time, account, easy, fix, card, phone, credit, log and password. 
This gives us a good idea of the types of topics we might find across the reviews.
For instance, log-in and password could be related to problems arising when users attempt to enter their password to log-in. 
Using TF-IDF, we get very similar results: easy ($225.93$), update ($221.34$), account ($195.56$), time ($193.83$), work ($189.18$), great ($187.86$), new ($149.05$), fix ($143.08$), need ($142.045$), love ($136.57$). 

\begin{figure}[htbp]
\centerline{\includegraphics[width=0.5\textwidth]{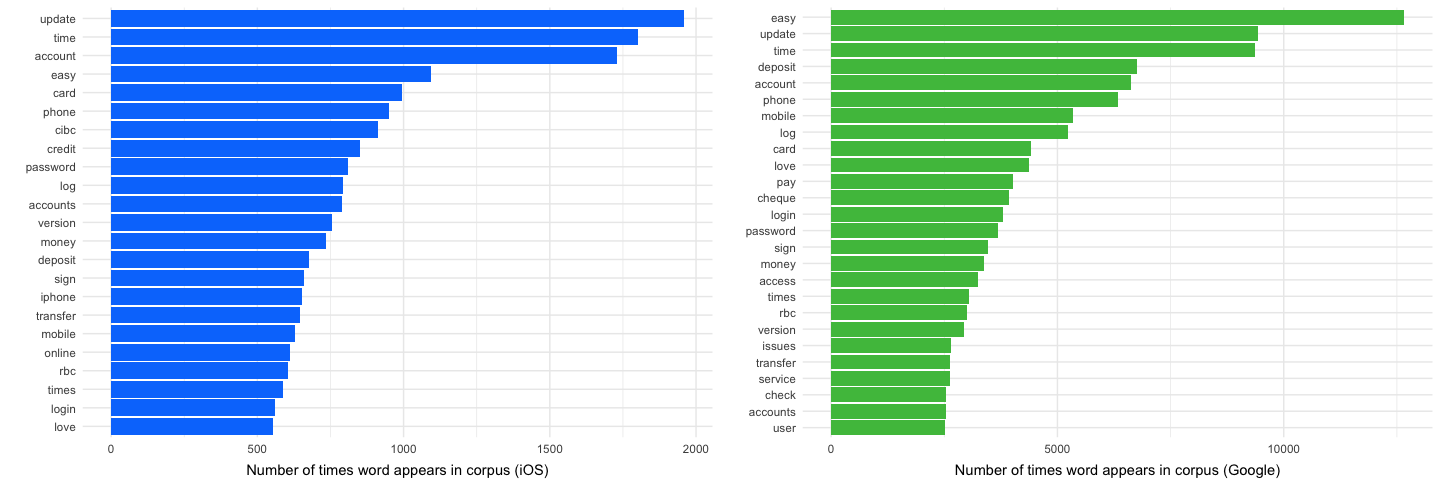}}
\caption{Top most frequent words in iOS and Google corpora by simple count}
\label{fig:top_words_1}
\end{figure}

``Fig.~\ref{fig:top_words_1}'' shows the top words found using a simple count mechanism for the iOS and Google corpora. 
From the top frequent words in ``Fig.~\ref{fig:top_words_1}'', it is evident that both iOS and Google users commonly discuss app updates, financial operations (e.g., deposits, transfers), account related (password, login) and overall performance. 
Key themes across both platforms include account management, secure access, and operational challenges. 
This provides a foundation for exploring topic modeling.

\begin{figure}[htbp]
\centerline{\includegraphics[width=0.5\textwidth]{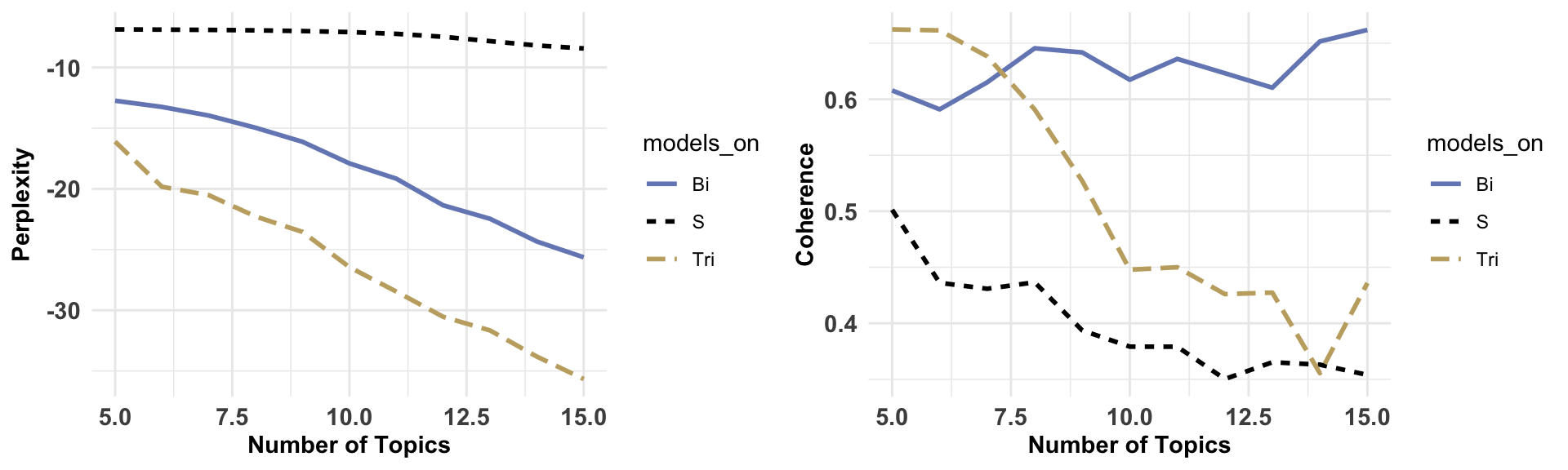}}
\caption{Model performance for different number of topics (iOS data). \ (left) Perplexity scores, where a lower perplexity score indicates a better model, (right) Coherence scores, where a higher coherence score indicates a better model.}
\label{fig:model_topic_comparisons_ios}
\end{figure}

For topic modeling, we first select the number of topics using perplexity and coherence. 
``Fig.~\ref{fig:model_topic_comparisons_ios}'' depicts the comparison of model performance based on the two scores for uni--grams, bi--grams and tri--grams on the iOS data. 
For perplexity, a smaller value indicates a better model. 
We see that for all $3$ n--grams, perplexity decreases as the number of topics increase. 
However, the rate of decrease is fastest in tri--grams, slightly slower in bi--grams and the slowest on uni--grams. 
Therefore, at least only according to perplexity, tri--grams produced better results than bi--grams which performed better than uni--grams. 
However, looking at the right hand panel in ``Fig.~\ref{fig:model_topic_comparisons_ios}'', we see that bi--grams perform the best. 
With coherence, a larger value is better. 
We see that for both uni--grams and tri--grams, increasing the number of topics decreases coherence. 
However, for bi--grams, this effect is the opposite and more topics slightly increase the coherence score. 
Based on this, the best model is trained on bi--gram corpus with $15$ topics ($p = -23.53, c = 0.60$) for iOS data. 

\begin{figure}[htbp]
\centerline{\includegraphics[width=0.45\textwidth]{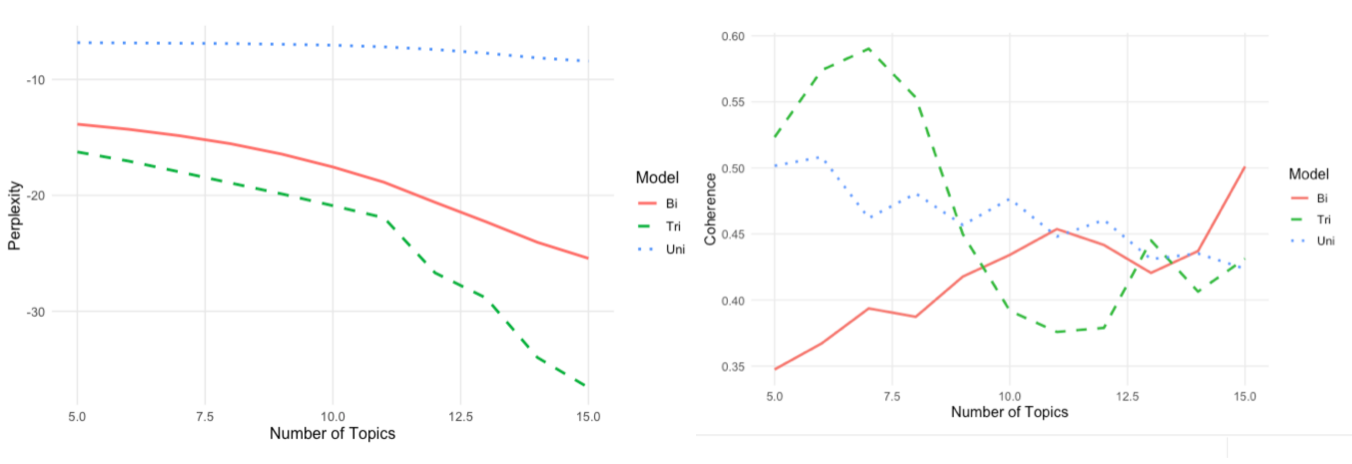}}
\caption{Model performance for different number of topics (Google data). \ (left) Perplexity scores, where a lower perplexity score indicates a better model, (right) Coherence scores, where a higher coherence score indicates a better model.}
\label{fig:model_topic_comparisons_google}
\end{figure}

From the left side of ``Fig.~\ref{fig:model_topic_comparisons_google}'' for Google data, we observe that as the number of topics increases, the perplexity consistently decreases across all three n--grams.
However, the rate of decrease varies.
Tri--grams show the steepest decline in perplexity, benefiting the most from an increased number of topics.
Bi-grams also improve as topics increase, though at a slightly slower rate, while uni--grams show the smallest improvement, with the slowest decline in perplexity.
From the right side, tri--grams show the fastest improvement in perplexity as the number of topics increases, followed by bi--grams with a slightly slower rate of improvement.
Uni--grams improve the least, with the slowest decline in perplexity.
From these results, we conclude that the bi--gram model with $15$ topics is the best overall for Google data. 
For the iOS data, the topics are as follows\footnote{Punctuation, full version of words and stop words are added for readability.}: 
\begin{itemize}
    \item Topic 1: don't know, account balance, user experience, home screen, thank (you) (so) much, enter password, update phone, time use(d), see balance, doesn't allow 
    \item Topic 2: uninstall reinstall, get error, never (a) problem, don't want, open account, last month, log back, longer use, need (to) use, click (function-name)
    \item Topic 3: face id, sign in, love app, can't use, first time, can't log, log account, can't see, use(d) (to) work, would (be) great
    \item Topic 4: user friendly, please try, try get(ting), older version, work anymore, much better, doesn't even, wait (in) line, don't use, people use
    \item Topic 5: credit card, doesn't work, say(s) (quote), fix this, use(d) it, I've use(d), please add, use(d) year, ever since, right now
    \item Topic 6: dark mode, latest update, much easier, use(d) time, 2 factor, factor authentication, noting work(s), seem(s) like, give(s) option(s), use(d) website
    \item Topic 7: please fix, pay bill(s), easy navigation, deposite check, last update, fix issue(s), debit card, since last, delete (and) reinstall, multiple time(s)
    \item Topic 8: new update, work(s) fine, can't even, tri(ed) later, mobile (app), waste (of) time, I've tri(ed)
    \item Topic 9: use(d) (app), access account(s), cheque deposit, work(s) well, mobile deposit, new phone, it (is) easy, check(ing) account, can't access, good job
    \item Topic 10: easy use, customer service, never work(s), every single, credit score, look(s) like, right away, call custom(er), tri(ed) login, feel(s) like
    \item Topic 11: time tri(ed), technical issue, tri(ed) login, every time, half (the) time, Apple pay, use (my) iPhone, can't login, trust device, face recognition 
    \item Topic 12: log in, great (app), would (be) nice, bill payment, error message, tri(ed) sign(ing), something went, went wrong, latest version, money (into/from) account
    \item Topic 13: can not, deposite cheque, won't let, works great, would like, fix it, even though, many time(s), stop(ped) work(ing), take picture
    \item Topic 14: every time, recent update, go(es) back, reset (my) password, online (banking), electronic transfer, bring back, chequing account, go (to) (the) branch, easy (and) convenient 
    \item Topic 15: step verification, 2 step, phone number, change (my) password, never (an) issue, two step, make sure, I've never, previous version, excellent service
\end{itemize}
Based on these $15$ topics, we note the following patterns: 
Consumers are discussing the \textbf{user experience} of the app, especially highlighting checking account balances, password entering and home screen. 
Issues and errors are brought up in several topics (Topic $2$, $3$, $5$, $11$, $12$). 
Specifically, face ID, log in, two factor authentication, Apple pay and technical issues. 
\textbf{Requests} are prevalent across the reviews with phrases such as "please add" requesting for features to be added to the apps. 
Several topics mention \textbf{updates} in various ways, but most seem to be about updates causing problems. 
Many people have pointed out uninstall reinstall not fixing their issues. 
Customer service, ease of use and e-transfers are brought up a few times. 

For Google, we identified several common topics, including login and account access issues, dissatisfaction with app updates and new designs, and technical problems following updates.
Other frequently mentioned topics highlight mobile cheque deposit issues, app stability concerns, and navigation challenges. 
Less frequent topics focus on specific issues such as fingerprint functionality, compatibility with Samsung Galaxy devices, security features, and card functionality.
These findings indicate that users face significant challenges with basic app functions and are frustrated by new updates that disrupt their experience.
While topics like bill payment features and customer service quality were mentioned, they were less prominent compared to broader concerns such as login issues and app stability. 
The topics are:
\begin{itemize}
    \item Topic 1: can’t even, don’t like, like new, login screen, paying bill(s), even log, fix problem, even get, money account, new interface
    \item Topic 2: technical issue(s), long time, waste time, never problem, update can’t, fix issue, several time(s), worked well, sd card, make sure
    \item Topic 3: deposit cheque, work(s) well, user friendly, deposit cheque, take picture, picture cheque, try deposit, cant deposit, check balance, don’t want
    \item Topic 4: keep(s) saying, ever since, can’t see, finger print, latest version, many time(s), much easier, I’m using, don’t understand, new feature(s)
    \item Topic 5:doesn’t work, cheque deposit, can’t get, new version, much better, still doesn’t, uninstall reinstall, another bank, mobile cheque, work anymore
    \item Topic 6: don’t know, uninstalled reinstalled, looks like, banking app(s), back old, need go, e transfer, bring back, even sign, check account
    \item Topic 7: please fix, won’t even, worked great, everytime try, fix asap, pretty good, debit card, even open, time go, I’ve used
    \item Topic 8: every time, time try, Samsung Galaxy, work(s) fine, would nice, used work, security question, previous version, every single, e transfer
    \item Topic 9: online banking, latest update, would like, can’t log, error message, like see, multiple time(s), uninstalling reinstalling, photo cheque, tried uninstalling
    \item Topic 10: won’t let, credit card, go back, let sign, let log, I’ve tried, play store, get work, google play, I’m happy
    \item Topic 11: new update, everything need, access account, never work(s), can’t access, easy navigate, user interface, account balance(s), account info, web browser
    \item Topic 12: old version, mobile banking, try later, transfer money, recent update, transfer fund(s), please add, please try, really like, hate new
    \item Topic 13: last update, since update, stopped working, new phone, since last, keep getting, worked fine, old one, need update, one star
    \item Topic 14: card number, work great, client card, great great, keeps crashing, remember card, love great, great love, enter card, don’t need
    \item Topic 15: pay bill(s), customer service, please update, bank account, night away, I’m sure, like able, one account, last week, every day
\end{itemize}

\subsection{Sentiment Analysis}
\label{sentimentanalysis}
For sentiment analysis, we apply the TextBlob function which calculates a polarity score for each review. 
We then convert the polarity scores into labels to compare with the manual labeling values for accuracy measures. 
TextBlob performs poorly on the iOS dataset with an accuracy score of only $24$\%.
Comparing the results of the manual label with TextBlob, we clearly see based on human language understanding that our manual labeling is far more accurate. 
Table \ref{tab: manual vs blob example} provides a few examples of manual labels and labels generated by TextBlob in the iOS dataset. 

\begin{table}[htbp]
    \caption{Examples of reviews with different labels provided by manual labeling and TextBlob (iOS Data)}
    \begin{center}
    \begin{tabular}{|p{1cm}|p{5cm}|p{0.5cm}|p{0.5cm}|}
    \hline
    Title & Review & Man$^1$ & Blob \\
    \hline
    Works terrible & Practically every time I open it, it asks me to verify and sends texts. asks if I like to stay logged in for future use with face id to avoid getting the text (i check the box to confirm this) doesn't matter anyways. & neg & neu \\
    \hline
    Not working in iOS$12.2$ & Won't work with ios$12.2$. Every time I open it, it turns white and stops working. I'm using the web version instead. & neg & neu \\
    \hline
    Horrible UI & They hired an intern for a new redesign? looks horrible & neg & neg \\
    \hline
    Crashes & The app crashes when I try to look at previous statements	on iphone mini & neg & pos \\ 
    \hline
    Solid & Solid app, easy to use. One thing I don't like: advertising lame products continually & pos & neu\\
    \hline
    \multicolumn{1}{l}{\emph{Notes:}} & \multicolumn{3}{l}{1: Manual labeling}\\
    \end{tabular}
    \label{tab: manual vs blob example}
    \end{center}
\end{table}
 
A similar procedure is followed for using NLTK's sentiment analyzer.
VADER's compound score is used for labeling. 
We select various values for the compound score threshold and find that we achieve the highest accuracy with compound threshold of $0.05$. 
However, the accuracy score is still very low at only $33.2\%$, doing only slightly better than TextBlob. 
For the example reviews in Table \ref{tab: manual vs blob example}, VADER labels are as follows: pos, neg, neg, neu, neu.

We train a multinomial Na\"ive Bayes classifier on the (iOS) data and achieve $67\%$ accuracy. 
This indicates that classification algorithms might perform much better in this instance. 
The LSTM introduced in section \ref{Data ana Methods} reports accuracy of $82\%$.
Since the LSTM has the highest accuracy score, it is used for sentiment analysis. 
We label the data, and split the original dataset into three separate datasets: only negative reviews ($5,475$ data points), only positive reviews ($3,413$) and only neutral reviews ($1,440$). 
This break-down also aligns with our expectations and discussion in section \ref{Data ana Methods}, especially regarding how the app with the lowest number of ratings had the highest number of reviews: people with extreme experiences are more likely to write reviews. 

For sentiment analysis on the Google data, we evaluated four methods: TextBlob, VADER, Multinomial Naive Bayes (MNB), and Long Short-Term Memory (LSTM).
Sentiment scores were classified into positive, neutral, and negative categories based on the mean and standard deviation (SD) of the scores.
The accuracy of each method was: TextBlob ($25.52$\%), VADER ($33.02$\%), MNB ($77.29$\%), and LSTM ($76.45$\%). 
MNB, with the highest accuracy, was selected as the optimal method. 
The model's predicted labels were compared with actual labels for the accuracy.
After classification, the sentiment predictions were divided into three subsets: $32,293$ negative reviews, $7,516$ neutral reviews, and $33,708$ positive reviews, which were prepared for further topic modeling analysis. 

\subsection{Topic Modeling - Sentiment Specific}
\label{Topic Modeling - Sentiment Specific}
Similar to section \ref{topicmodeling}, we do topic modeling for each of the positive, negative and neutral datasets. 
That is, for each dataset, we find the best number of topics for each n--gram model, and finally decide the best overall model. 
To avoid repetition, we eliminate the repeat of the detail breakdown of topics for the positive and negative subsets of the datasets in this subsection, but rather discuss the final outcomes more generally. 
The results for the topic modeling models on n--grams for all three datasets on the iOS data are summarized in Table \ref{tab:topicmodelingaftersentiment}.

For negative reviews, the bi--gram model has the best $p$ score but the tri--gram model has the best $c$ score. 
To decide which model is chosen as the final best, we calculate a score value that is the weighted sum of $p$ and $c$ as follows, giving both metrics equal importance:
\begin{equation}
    \label{scoreeq}
    score = 0.5 \times (1 - p) + 0.5 \times c 
\end{equation}
The score for the bi--gram model is $12.80$ which is higher than the score for the tri--gram based model ($11.39$). 
Therefore, the best model is the bi--gram model with $15$ topics (for negative iOS reviews).
Similarly, for both neutral and positive reviews, the tri--gram models are the best, with $15$ and $13$ topics, respectively. 

\begin{table}[htbp]
    \caption{Comparison of topic modeling results (iOS Data)}
    \begin{center}
    \begin{tabular}{cccccc}
    \hline
    n-gram & S$^1$ & $p^2$ & $c^3$ & $|T|$ & score\\
    \hline
    uni & neg & $-9.536$  &  $0.389$ & $15$  & --- \\
    bi  & neg &  $-23.903$ &  $0.703$ &  $15$ & $12.80$ \\
    tri & neg &  $-21.000$ &  $0.779$ &  $7$ & $11.39$ \\
    \hline
    uni & neu & $-9.076$ & $0.355$  &  $15$ & --- \\
    bi  & neu & $-20.940$ & $0.730$  &  $15$ & $11.34$ \\
    tri & neu & $-29.120$ & $0.569$  &  $15$ & $15.34$ \\
    \hline 
    uni & pos & $-9.379$ & $0.371$  & $15$ & --- \\
    bi  & pos & $-23.302$ & $0.724$  & $15$  & $12.51$ \\
    tri & pos & $-30.647$ & $0.686$  & $13$  & $16.17$\\
    \hline
    \multicolumn{1}{l}{\emph{Notes:}} & \multicolumn{5}{l}{1: Sentiment \ \ \ 2: Perplexity \ \ \ 3: Coherence}\\
    \multicolumn{6}{l}{Score is not calculated for the worst performing topic model.}
    \end{tabular}
    \label{tab:topicmodelingaftersentiment}
    \end{center}
\end{table}

For the Google data, the negative dataset had the highest coherence score of $13.65$ with $15$ topics bi-grams, while the neutral dataset also performed best with $15$ topics bi-grams, getting a coherence score of $13.29$.
Trigrams with $14$ topics yielded the highest coherence score of $18.03$ for the positive dataset.
Since we focus on the positive and negative datasets, the neutral dataset is excluded from further analysis. 

\begin{table}[htbp]
    \caption{Comparison of topic modeling results (Google Data)}
    \begin{center}
    \begin{tabular}{cccccc}
    \hline
    n-gram & S$^1$ & $p^2$ & $c^3$ & $|T|$ & score\\
    \hline
    uni & neg & $-6.884$ &  $0.474$ & 5 & --- \\
    bi  & neg & $-25.687$ &  $0.610$ & 15 & 13.65 \\
    tri & neg & $-18.014$ &  $0.644$ & 5 & 9.83 \\
    \hline
    uni & neu & $-8.119$ &  $0.363$ & 14 & --- \\
    bi  & neu & $-24.891$ &  $0.679$ & 15 & $^*$ \\
    tri & neu & $-19.002$ &  $0.630$ & 7 & --- \\
    \hline
    uni & pos & $-6.856$ &  $0.500$ & 5 & --- \\
    bi  & pos & $-24.796$ &  $0.612$ & 15 & 13.20 \\
    tri & pos & $-34.593$ &  $0.464$ & 14 & 18.03\\
    \hline
    \multicolumn{1}{l}{\emph{Notes:}} & \multicolumn{5}{l}{1: Sentiment \ \ \ 2: Perplexity \ \ \ 3: Coherence}\\
    \multicolumn{6}{l}{*: Both $p$ and $c$ are best.}\\
    \multicolumn{6}{l}{Score is not calculated for the worst performing topic model.}\\
    \end{tabular}
    \label{tab:topicmodelingaftersentiment_google}
    \end{center}
\end{table}

We run the chosen topic models, omitting the neutral group as it is not useful for our analysis.
We will not discuss the topics in minute detail, rather present the general trends in each group. 
For easier compartmentalization, we present the results for the negative subset of reviews first and positive subset second for bother datasets. 

\subsubsection{Negative Reviews Results}
For negative sentiment group in the iOS data, the topics are as follows:
\begin{itemize}
    \item \textbf{User Experience} with specific instances of "keeps crashing" and "previous versions"
    \item \textbf{Issues} with specific problems such as "takes forever" indicating speed as a problem area
    \item \textbf{Problematic Update} indicating the app used to work ("worked great before" or "since the update") but the update has caused problems or "bring back [functionality]" showing that some removal is not appreciated  
    \item \textbf{Login Issues} specific log in issues such as 2 step verification, and slowness ("keeps loading")
    \item \textbf{Technical Issues} specifically mentioned "facial recognition" and "login screen", "touch id", "password" (all of which are somewhat related to login issues as well)
    \item \textbf{Customer Service} with more extreme vocabulary such as "waste of time" 
\end{itemize}
As is apparent, the negative reviews focus more on technical issues with the apps. 
Most topics are related to sudden onset of problems after an update for the app. 
\textbf{Updates} are usually rolled out to fix bugs and address issues in prior versions, however, it seems like updates cause new challenges. 
This is an interesting area observation which is a worthwhile research avenue to see why it is that updates seem to cause problems for some. 
\textbf{Speed} of the apps is another important aspect-based topic which is brought up in negative reviews. 
While speed is a subjective notion (i.e., $5$ seconds may be fast to one person but slow to another), in our modern world, an app taking more than a few short seconds to load or do anything is not acceptable. 

Banks should address issues regarding \textbf{login} problems, especially related to password management and 2-factor authentication and bio-metrics. 
These problems could be caused by \textbf{updates}, which is another area banks need to improve upon. 
Perhaps more rigorous quality management or testing before rolling out updates, not removing features that were frequently used by users, and not changing too much of the interface are good strategies for banks to follow. 
Lastly, banks must prioritize resolving disputes with \textbf{customer service}, especially addressing the significant delays customers face when trying to reach support.
Addressing these concerns would greatly improve customer satisfaction and customer retention for the banks \cite{ahmad17}. 

The key topics from all Google reviews show that users face significant challenges with \textbf{basic app functions} and frustration with \textbf{new app versions}. 
The bi-gram analysis of the negative dataset reinforces these findings, highlighting recurring issues such as \textit{login difficulties, frequent crashes, slow performance}, and \textit{dissatisfaction with updates}.
Additional concerns include problems with specific features like mobile deposits, bill payments, e-transfers, and credit card management.
Interestingly, the negative dataset reveals a new concern: many users express \textbf{dissatisfaction with customer service} and indicate they are considering switching to other banks. 

\subsubsection{Positive Reviews Results}
For the iOS dataset, the positive reviews were far fewer than negative reviews and the overall topics were heavily dominated by negative sentiments.
However, we see some optimistic patterns in the positive reviews subset, indicated as follows: 
\begin{itemize}
    \item \textbf{Easy to Use} mentioned in various forms ("great easy (to) use", "it's easy (to) navigate", "it's easy (to) use", "super easy (to) use", "clean (and) easy (to) use") 
    \item \textbf{Overall satisfaction, but suggestions for further improvements} it seems that many of the positive reviews also give suggestions to banks ("would also (be) nice", "please change back", "nice (to be) able (to) transfer", "would like (to) see", "would (be) (a) nice option")
    \item \textbf{Update} surprisingly, not all updates are bad ("love new update") 
    \item \textbf{Customer Service} "best customer service", "great customer service" seems to have a very high correlation with higher ratings
    \item \textbf{Specific bonus features} such as "free FICO score check" or "helps build budget" or "spend analyzer easily", and "analyzer easily track"
    \item \textbf{User Friendly} and convenient 
\end{itemize}
It appears that most satisfied customers are happy with an app that is \textbf{easy to use}.
This aligns with various technology adoption models and theories such as Technology Adoption Model (TAM) \cite{davis89}.
Additionally, many of the positive reviews seem to include \textbf{suggestions} for even \textbf{further improvements} to the app.
Banks should leverage these reviews and take inspiration from customer suggestions. 
\textbf{Updates} are surprisingly present in positive reviews as well, indicating not all updates cause issues. 
Similar to updates, \textbf{customer service} is also present in the positive subset of reviews, indicating a satisfactory with customer service (solved the issue, was quick and professional, addressed all the concerns of the customer, etc) is highly important. 

Few \textbf{specific features} are mentioned across the positive review subset in the iOS data.
These are app or bank specific features, and their additions seem to indicate a positive feedback from customers. 
For example, tools that help customers \textit{budget} or analyze and track their spending seem to be popular with customers. 
More banks should integrate such features into their apps to educate, aid and inform their customers. 
Other features mentioned are \textbf{free credit score checks}, which again more banks should offer to customers within the mobile app. 
Lastly, somewhat related to ease of use, having a \textbf{user-friendly} app (be it design/interface or functionality) is most definitely needed to improve customer experience and satisfaction. 

Since topics from all Google reviews were focused on issues and challenges, this project also explored what the m-banking apps do well.
The analysis of the positive dataset on Google highlights key strengths, particularly in \textbf{customer service} and \textbf{usability}.
Users frequently praise the app for being \textbf{user-friendly}, \textbf{easy to navigate}, and \textbf{effective} for everyday banking tasks.
\textbf{Reliable performance} is another common topic, with users commending the app’s \textbf{stability} in handling bill payments, fund transfers, and account management. 
Also, users appreciate \textbf{convenient} \textbf{features} like bill payments, direct deposits, and \textbf{e-transfers}, noting that these functionalities \textit{save time} and make banking more efficient.

\section{Conclusions}
\label{Conclusions}
This research provides insights into the experiences and concerns of users interacting with mobile banking (m-banking) applications (apps) on iOS and Android (Google Play) platforms. 
Through topic modeling, sentiment analysis, and classification, several key findings emerged.

By performing topic modeling before sentiment analysis, we gained an initial understanding of all reviews extracted from Google Play and iOS App stores.
This step allowed us to identify overarching topics and establish a clear direction for further analysis.
The topic modeling analysis revealed common themes across user reviews.
On iOS, frequent topics included login issues, password entry challenges, problems with app updates, and requests for additions of specific features.
Similarly, for Android, issues like account access, cheque deposit problems, technical glitches, and dissatisfaction with updates were prevalent. 
These findings suggest that users frequently face significant barriers when performing basic tasks, which often contribute to their dissatisfaction.
Bi--grams emerged as the most effective model for capturing coherent topics across both datasets, indicating that phrases provide richer context than single words or trigrams in this topic modeling instance.

Following this, sentiment analysis was performed which highlighted limitations in current tools for assessing user sentiments.
While tools like TextBlob and VADER showed low accuracy scores ($24$\% and $33.2$\% respectively) on iOS data and not much better on Google data ($25.52\%, 33.02$\%).
Manual labeling proved more reliable in capturing nuanced sentiments.
A Na\"ive Bayes classifier, trained on labeled data, achieved an improved accuracy of $67$\%, demonstrating the potential for machine learning models to enhance sentiment classification in this domain.
For the iOS dataset, a Recurrent Neural Network (LSTM) was used to classify the dataset into positive, negative and neutral groups. 
For the Google dataset, the Na\"ive Bayes classifier was used for the same task. 

Following this, each subset of the data (positive, negative and neutral) was used for another set of topic modeling to discover sentiment-based topics. 
For the negative datasets, the results further reinforced the topics identified before, confirming recurring topics such as issues like login difficulties, and crashes, dissatisfaction with updates, as well as customer service delays. 
In contrast, the results from the positive dataset offered more nuanced insights, showcasing what these mobile banking apps are doing well. 

Users frequently praised features like user-friendly interfaces, reliable performance in tasks as well as time-saving and unique functionalities.
Ease of use emerged as the most important feature in indicating positive experience for customers. 
These findings provide a balanced view of both the challenges and strengths of mobile banking apps, offering valuable insights for app improvement and user satisfaction. 

Overall, this study underscores the need for m-banking app developers to address recurring user concerns, particularly login-related issues, update stability, and feature requests. 
Additionally, advancements in sentiment analysis tools and the integration of machine learning models can provide more accurate insights into user feedback, enabling data-driven app improvements.
Future work could 1) explore more sophisticated modeling techniques, 2) include other languages in the reviews, 3) consider emoji's and sarcasm detection, and 4) extend the analysis to other apps and even app categories to compare the results.
The last point would make for an interesting research topic where the researchers could address if the specific complaints and praises are related to the apps being m-banking apps or whether these are general across all or most other categories. 
Another interesting avenue of research is to compare the differences between the two OS platforms (iOS and Android). 
Lastly, future research could consider a bot (fake review) detection to eliminate the possibility of inflated reviews.

\section*{Acknowledgment}
The authors acknowledge the use of ChatGPT4 (https://chatgpt.com) for improving and shortening the writing in this manuscript. 
Originally written text by the authors were given to the LLM (paragraph by paragraph) where the prompt asked to keep the language and all the points, but shorten the text to 60\% of the original size. 
Yekta was responsible for all analysis related to the iOS dataset, as well as writing the introduction, methodology and results for the iOS portion of analysis. 
Kisty was responsible for all analysis related to the Google play dataset, conclusion and all results for the Google portion of analysis. 


\vspace{12pt}

\end{document}